\pdfoutput=1

\documentclass[11pt]{article}

\usepackage[]{emnlp2021}

\usepackage{times}
\usepackage{latexsym}

\usepackage[T1]{fontenc}

\usepackage[utf8]{inputenc}

\usepackage{microtype}

\usepackage{adjustbox}
\usepackage{tikz}
\usepackage{tkz-graph}
\usetikzlibrary{arrows,calc,matrix,trees,patterns,automata,positioning,shapes.geometric}
\usepackage{scrextend}
\usepackage{chngcntr}
\usepackage[frozencache=true,cachedir=minted-cache]{minted} 
\setminted{style=vs}
\usepackage{graphicx} 
\usepackage{booktabs} 
\usepackage{xparse}   
\usepackage{wrapfig}
\usepackage{multirow}
\usepackage{tikz-dependency}
\usepackage{comment}
\usepackage{pgfplots}
\pgfplotsset{compat=1.9}

\newcommand{\compacturl}[1]{{\fontsize{10}{8.5}\selectfont\textls*[-70]{\url{#1}}}}
\newcommand{\compacturll}[1]{{\fontsize{4}{3}\selectfont\textls*[-100]{\url{#1}}}}

\title{{COMBO}: {S}tate-of-the-{A}rt {M}orphosyntactic {A}nalysis}

\author{Mateusz Klimaszewski$^{1,2}$ Alina Wr\'{o}blewska$^2$  \\
$^1$Warsaw University of Technology \\ 
$^2$Institute of Computer Science, Polish Academy of Sciences \\
{\tt m.klimaszewski@ii.pw.edu.pl}\\{\tt alina@ipipan.waw.pl}
}

\date{}

\begin{document}
\maketitle
\begin{abstract}
We introduce COMBO -- a~fully neural NLP~system for accurate part-of-speech tagging, morphologi\-cal analysis, lemmatisation, and (enhanced) dependency parsing. It predicts categorical morphosyntactic features whilst also exposes their vector representations, extracted from hidden layers. COMBO is an~easy to install Python package with automatically downloadable pre-trained models for over 40 languages.
It maintains a balance between efficiency and quality. As it is an~end-to-end system and its modules are jointly trained, its training is competitively fast. As its models are optimised for accuracy, they achieve often better prediction quality than SOTA. 
The~COMBO library is available at: \url{{https://gitlab.clarin-pl.eu/syntactic-tools/combo}}.
\end{abstract}

\section{Introduction}
\begin{figure*}[ht!]
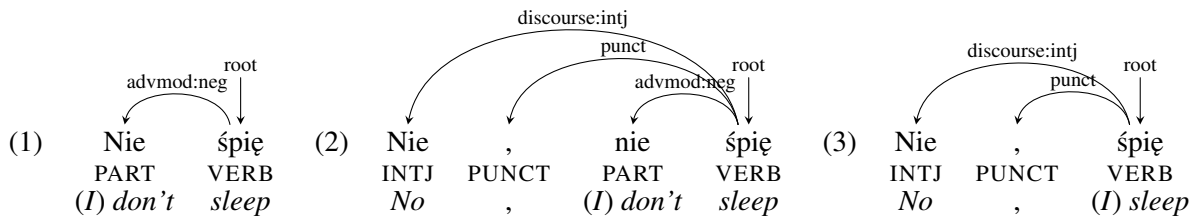

\centering
\begin{dependency}[theme = simple, label style={font=\normalsize}]
   \begin{deptext}[column sep=0.25cm]
     (1) \& Nie \& śpię \\
       \& \textsc{part} \& \textsc{verb}\\
       \& (\textit{I}) \textit{don't} \& \textit{sleep}\\
   \end{deptext}
   \depedge{3}{2}{advmod:neg}
   \deproot[edge height=5ex]{3}{root}
\end{dependency}
\begin{dependency}[theme = simple, label style={font=\normalsize}]
   \begin{deptext}[column sep=0.25cm]
      (2) \& Nie \& , \& nie \& śpię \\
      \&\textsc{intj} \& \textsc{punct} \& \textsc{part} \& \textsc{verb}\\
       \& \textit{No} \& , \& (\textit{I}) \textit{don't} \& \textit{sleep}\\
   \end{deptext}
   \depedge{5}{4}{advmod:neg}
   \deproot[edge height=5ex]{5}{root}
   \depedge{5}{3}{punct}
   \depedge{5}{2}{discourse:intj}
\end{dependency}
\begin{dependency}[theme = simple, label style={font=\normalsize}]
   \begin{deptext}[column sep=0.25cm]
      (3) \& Nie \& , \& śpię \\
      \& \textsc{intj} \& \textsc{punct} \& \textsc{verb}\\
      \& \textit{No} \& , \& (\textit{I}) \textit{sleep}\\
   \end{deptext}
   \deproot[edge height=5ex]{4}{root}
   \depedge{4}{3}{punct}
   \depedge{4}{2}{discourse:intj}
\end{dependency}
\caption{UD trees of Polish sentences: (1) and (2) mean a~\textit{non-sleeping} situation and (3) means \textit{sleeping}.}
\label{fig:trees}
\end{figure*}

Natural language processing (NLP) has long recognised morphosyntactic features as necessary for solving advanced natural language understanding (NLU) tasks. An~enormous impact of contextual language models on presumably all NLP tasks has slightly weakened the~importance of morphosyntactic analysis. As morphosytnactic features are encoded to some extent in contextual word embeddings \cite[e.g.][]{tenney2018what,lin-etal-2019-open}, doubts arise as to whether explicit morphosyntactic knowledge is still needed. For example, \citet{glavas-vulic-2021-supervised} have recently investigated an~intermediate fine-tuning contextual language models on the~dependency parsing task and suggested that this step does not significantly contribute to advance NLU models. Conversely, \citet{warstadt-etal-2019-investigating} reveal the~powerlessness of contextual language models in encoding linguistic phenomena like negation. This is in line with our intuition about representing negation in Polish sentences (see Figure \ref{fig:trees}). It does not seem trivial to differentiate between the~contradicting meanings of these sentences using contextual language models, as the~word context is similar. The~morphosyntactic features, e.g. parts of speech PART vs. INTJ, and dependency labels \textit{advmod:neg} vs. \textit{discourse:intj}, could be beneficial in determining correct reading.

In order to verify the~influence of explicit morphosyntactic knowledge on NLU tasks, it is necessary to design a~technique for injecting this knowledge into models or to build morphosyntax-aware representations. The~first research direction was initiated by \citet{glavas-vulic-2021-supervised}. Our objective is to provide a~tool for predicting high-quality morphosyntactic features and exposing their embeddings. These~vectors can be directly combined with contextual word embeddings to build morphosyntactically informed word representations. 

The emergence of publicly available NLP datasets, e.g. Universal Dependencies \cite{ud25data}, stimulates the development of NLP systems. Some of them are optimised for efficiency, e.g. spaCy \cite{spacy}, and other for accuracy, e.g. UDPipe \cite{straka-2018-udpipe}, the~Stanford system \cite{dozat-manning-2018-simpler}, Stanza \cite{qi-etal-2020-stanza}. In this paper, we introduce COMBO, an~open-source fully neural NLP system which is optimised for both training efficiency and prediction quality. Due to its~end-to-end architecture, which is an~innovation within morphosyntactic analysers, COMBO is faster in training than the~SOTA pipeline-based systems, e.g. Stanza. As a result of applying modern NLP solutions (e.g. contextualised word embeddings), it qualitatively outperforms other systems.

COMBO analyses tokenised sentences and predicts morphosyntactic features of tokens (i.e. parts of speech, morphological features, and lemmata) and syntactic structures of sentences (i.e. dependency trees and enhanced dependency graphs). At the same time, its module, COMBO-vectoriser, extracts vector representations of the~predicted features from hidden layers of individual predictors. COMBO user guide is in \S \ref{sec:getting_started} and a~live demo is available on the~website \url{http://combo-demo.nlp.ipipan.waw.pl}. 

\paragraph{Contributions} 1) We implement COMBO (\S \ref{sec:architecture}), a~fully neural NLP system for part-of-speech tagging, morphological analysis, lemmatisation, and (enhanced) dependency parsing, together with COMBO-vectoriser for revealing vector representations of predicted categorical features. COMBO is implemented as a~Python package which is easy to install and to integrate into a~Python code. 
2) We pre-train models for over 40 languages that can be automatically downloaded and directly used to process new texts. 
3) We evaluate COMBO and compare its performance with two state-of-the-art systems, spaCy and Stanza (\S \ref{sec:evaluation}).

\section{COMBO Architecture}
\label{sec:architecture}
\begin{figure}[!h]
\centering
\includegraphics[width=\linewidth]{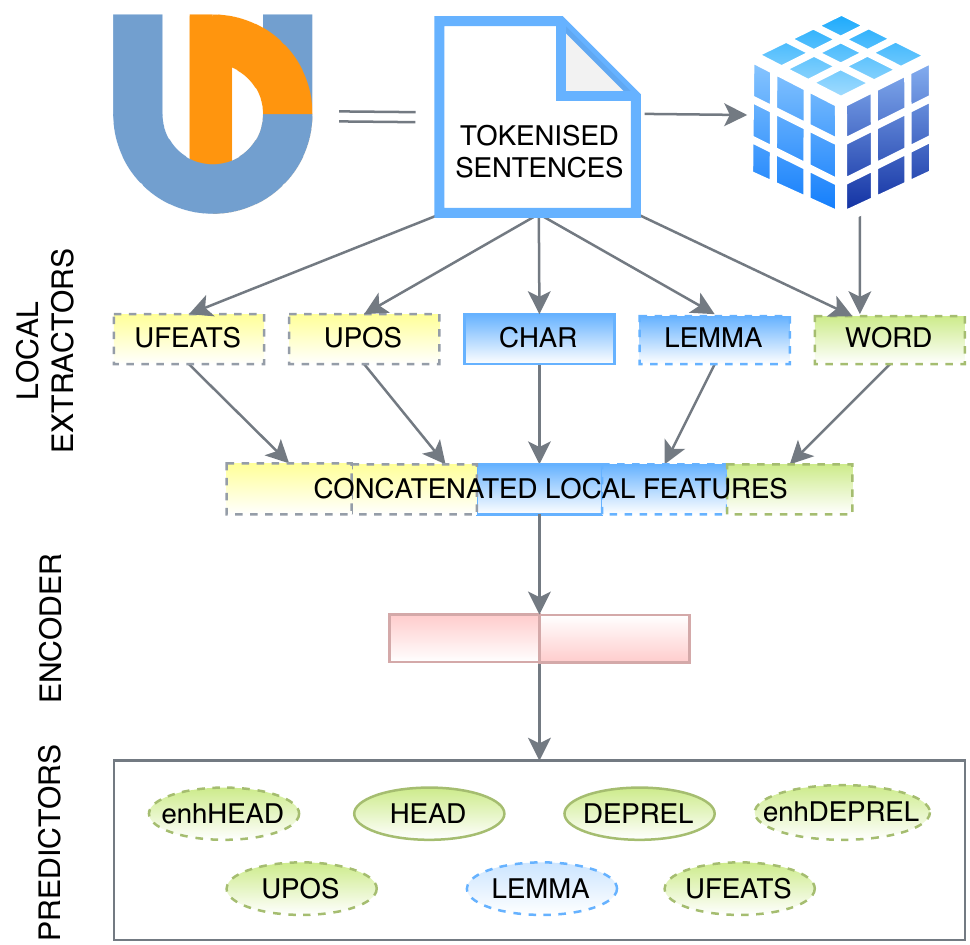}
\caption{\label{fig:overview}COMBO architecture. Explanations:\\ \includegraphics[scale=0.8]{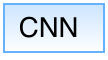}
\includegraphics[scale=0.8]{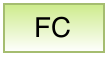}
\includegraphics[scale=0.8]{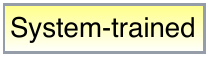}
\includegraphics[scale=0.8]{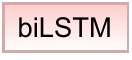}
\includegraphics[scale=0.8]{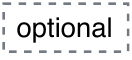}
\includegraphics[scale=0.8]{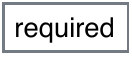}}
\end{figure}

COMBO's architecture (see Figure~\ref{fig:overview}) is based on 
the~forerunner \cite{rybak-wroblewska-2018-semi} implemented in the~Keras framework. 
Apart from a~new implementation in the~PyTorch library \cite{NEURIPS2019_9015}, the~novelties are the~BERT-based encoder, the~EUD prediction module, and COMBO-vectoriser extracting embeddings of UPOS and DEPREL from the~last hidden layers of COMBO's tagging and dependency parsing module, respectively. This section provides an~overview of COMBO's modules. Implementation details are in Appendix \ref{sec:implementationDetails}.

\paragraph{Local Feature Extractors}
\label{sec:localFeatureExtraction}
Local feature extractors (see Figure~\ref{fig:overview}) 
encode categorical features (i.e. words, parts of speech, morphological features, lemmata) into vectors. The~feature bundle is configurable and limited by the~requirements set for COMBO. For instance, if we train only a~dependency parser, the~following features can be input to COMBO: internal character-based word embeddings (\textsc{char}), pre-trained word embeddings (\textsc{word}), and embeddings of lemmata (\textsc{lemma}), parts of speech (\textsc{upos}) and morphological features (\textsc{ufeats}). If we train a~morphosyntactic analyser (i.e. tagger, lemmatiser and parser), internal word embeddings (\textsc{char}) and pre-trained word embeddings 
(\textsc{word}), if available, are input to COMBO.

Words and lemmata are always encoded using character-based word embeddings (\textsc{char} and \textsc{lemma}) estimated during system training with a~dilated convolutional neural network (CNN) encoder \cite{dcnn:2015,strubell-etal-2017-fast}. 

Additionally, words can be represented using pre-trained word embeddings (\textsc{word}), e.g. fastText \cite{grave2018learning}, or BERT \cite{bert:2018}. The~use of pre-trained embeddings is an~optional functionality of the system configuration.
COMBO freezes pre-trained embeddings
(i.e. no fine-tuning) and uses their transformations, i.e. embeddings are transformed by a~single fully connected (FC) layer.

Part-of-speech and morphological embeddings (\textsc{upos} and \textsc{ufeats}) are estimated during system training. 
Since more than one morphological feature can attribute a~word, embeddings of all possible features are estimated and averaged to build a~final morphological representation.

\begin{figure}[h!]
    \centering
    \includegraphics[width=0.5\linewidth]{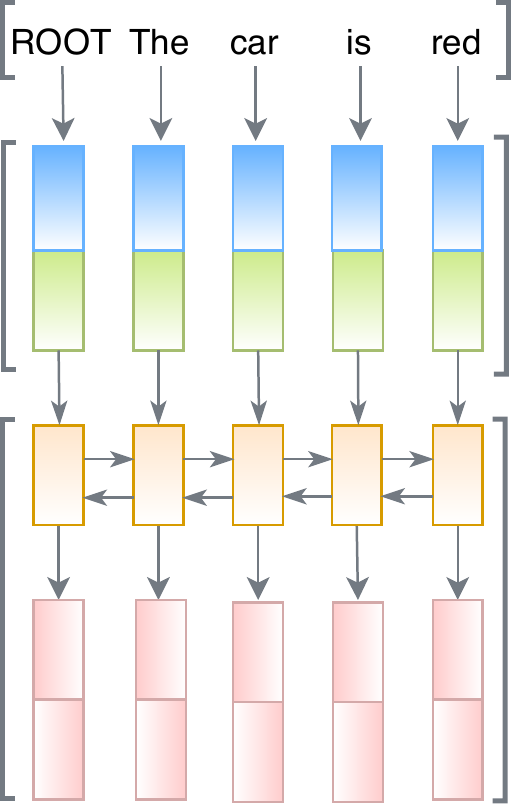}
    \caption{\label{fig:global}Estimation of global feature vectors.\\ \includegraphics[scale=0.8]{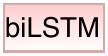} \includegraphics[scale=0.8]{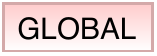}}
\end{figure}

\paragraph{Global Feature Encoder}
\label{sec:globalFeatureExtraction}
The~encoder uses concatenations of local feature embeddings. A sequence of these vectors representing all the~words in a~sentence is processed by a~bidirectional LSTM \cite{lstm:1997,bilstm:2005}. The~network learns the~context of each word and encodes its global (contextualised) features (see Figure \ref{fig:global}). Global feature embeddings are input to the~prediction modules.

\vspace*{-6mm}

\paragraph{Tagging Module}
\label{sec:tagger}
The~tagger takes global feature vectors as input and predicts a~universal part of speech (\textsc{upos}), a~language-specific tag (\textsc{xpos}), and morphological features (\textsc{ufeats}) for each word. The~tagger consists of two linear layers followed by a~softmax. Morphological features build a~disordered set of category-value pairs (e.g. Number=Plur). Morphological feature prediction is thus implemented as several classification problems. The~value of each morphological category is predicted with a~FC network. Different parts of speech are assigned different sets of morphological categories (e.g. a~noun can be attributed with grammatical gender, but not with grammatical tense). The~set of possible values is thus extended with the~NA (not applicable) symbol. It allows the model to learn that a~particular category is not a property of a word.

\paragraph{Lemmatisation Module}
\label{sec:lemmatizer}
The~lemmatiser uses an~approach similar to character-based word embedding estimation. A~character embedding is concatenated with the~global feature vector and transformed by a linear layer. The lemmatiser takes a sequence of such character representations and transforms it using a dilated CNN. The softmax function over the result produces the sequence of probabilities over a character vocabulary to form a lemma.
\begin{figure}[!h]
\centering
\includegraphics[width=1\linewidth]{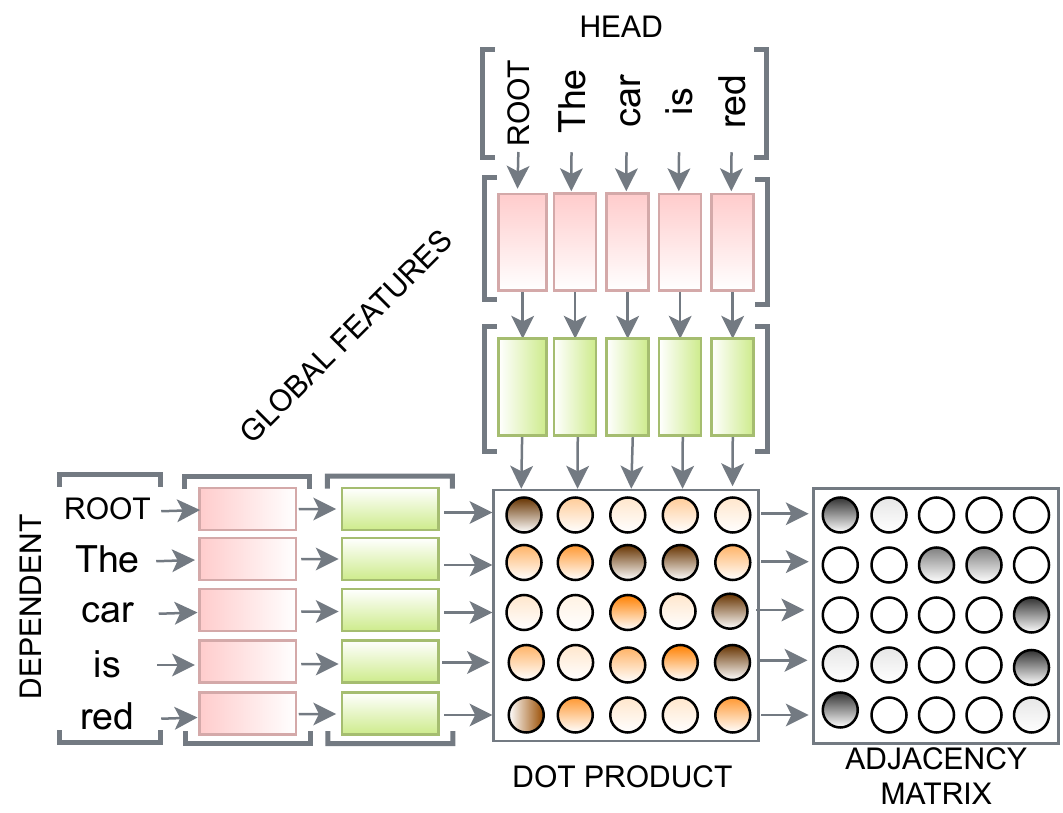}
\caption{\label{fig:tree}Prediction of dependency arcs.}
\end{figure}

\paragraph{Parsing Module}
\label{sec:parser}
Two single FC layers transform global feature vectors into head and dependent embeddings (see Figure \ref{fig:tree}). Based on these representations, a~dependency graph is defined as an~adjacency matrix with columns and rows corresponding to heads and dependents, respectively. The~adjacency matrix elements are dot products of all pairs of the~head and dependent embeddings (the~dot product determines the certainty of the~edge between two words). The~softmax function applied to each row of the~matrix predicts the~adjacent head-dependent pairs. This approach, however, does not guarantee that the~resulting adjacency matrix is a~properly built dependency tree. The~Chu-Liu-Edmonds algorithm \cite{chuLiu:65,Edmonds:67} is thus applied in the~last prediction step.
\begin{figure}[h!]
\centering
\includegraphics[width=\linewidth]{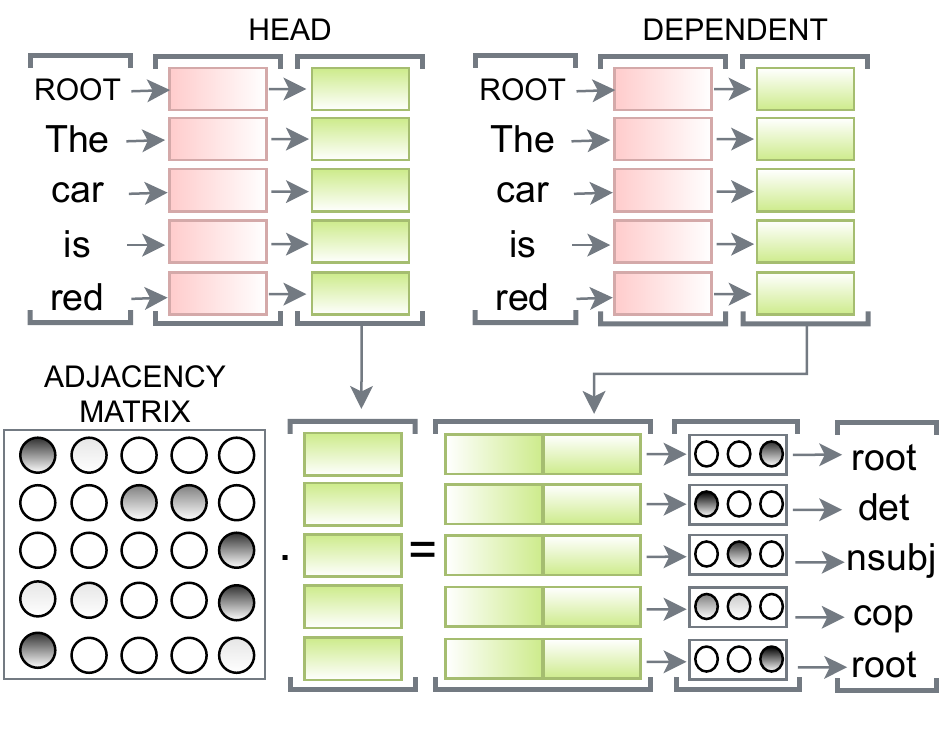}
\caption{\label{fig:labels}Prediction of grammatical functions.}
\end{figure}

The~procedure of predicting words' grammatical functions (aka dependency labels) is shown in Figure \ref{fig:labels}.
A~dependent and its head are represented as vectors by two single FC layers. 
The~dependent embedding is concatenated with the~weighted average of (hypothetical) head embeddings. The~weights are the values from the corresponding row of the~adjacency matrix, estimated by the~arc prediction module. Concatenated vector representations are then fed to a~FC layer with the~softmax activation function to predict dependency labels.

\paragraph{EUD Parsing Module}
Enhanced Universal Dependencies (EUD) are predicted similarly to dependency trees. The~EUD parsing module is described in details in \citet{klimaszewski-wroblewska-2021-combo}.

\begin{table*}[ht!]
\renewcommand\tabcolsep{8pt}
\setlength\aboverulesep{1pt}
\setlength\belowrulesep{1pt}
\fontsize{10}{12}\selectfont{
\begin{center}
\begin{tabular}{lrrrrrrrrr}
\toprule
System & UPOS & XPOS & UFeat & Lemma & UAS & LAS & CLAS & MLAS & BLEX \\
\midrule
\multicolumn{10}{c}{English EWT (isolating)} \\
\hline
spaCy
&93.79
&93.10
&94.89
&NA
&83.38
&79.76
&75.74
&68.91
&NA \\
Stanza
&96.36
&96.15
&97.01
&\bf98.18
&89.64
&86.89
&83.84
&79.44
&82.03\\
COMBO
&95.60
&95.21
&96.60
&97.43
&88.56
&85.58
&82.35
&76.56
&79.78 \\
COMBO$_{\textsc{bert}}$ 
&\bf96.57
&\bf96.44
&\bf97.24
&97.86
&\bf91.76
&\bf89.28
&\bf86.83
&\bf81.71
&\bf84.38 \\
\hline
\multicolumn{10}{c}{Arabic PADT (fusional)}  \\
\hline
spaCy
&90.27
&82.15
&82.70
&NA
&74.24
&67.28
&63.28
&50.48
&NA\\
Stanza
&96.98
&93.97
&94.08
&\bf95.26
&87.96
&83.74
&80.57
&74.96
&\bf76.80\\
COMBO
&96.71
&93.72
&93.83
&93.54
&87.06
&82.70
&79.46
&73.25
&73.64\\
COMBO$_{\textsc{bert}}$
&\bf97.04
&\bf94.83
&\bf95.05
&93.95
&\bf89.21
&\bf85.09
&\bf82.36
&\bf76.82
&76.67\\
\hline
\multicolumn{10}{c}{Polish PDB (fusional)} \\
\hline
spaCy
&96.14
&86.94
&87.41
&NA
&86.73
&82.06
&79.00
&65.42
&NA \\
Stanza
&98.47
&94.20
&94.42
&97.43
&93.15
&90.84
&88.73
&81.98
&85.75 \\
COMBO
&98.24
&94.26
&94.53
&97.47
&92.87
&90.45
&88.07
&81.31
&85.53\\
COMBO$_{\textsc{bert}}$
&\bf98.97
&\bf96.54
&\bf96.80
&\bf98.06
&\bf95.60
&\bf93.93
&\bf92.34
&\bf87.59
&\bf89.91 \\
\hline 
\multicolumn{10}{c}{Finnish TDT (agglutinative)}  \\
\hline
spaCy
&92.15
&93.34
&87.89
&NA
&80.06
&74.75
&71.52
&61.95
&NA\\
Stanza
&97.24
&97.96
&95.58
&\bf95.24
&89.57
&87.14
&85.52
&80.52
&\bf81.05\\
COMBO
&96.72
&98.02
&94.04
&88.73
&89.73
&86.70
&84.56
&77.63
&72.42\\
COMBO$_{\textsc{bert}}$
&\bf98.29
&\bf99.00
&\bf97.30
&89.48
&\bf94.11
&\bf92.52
&\bf91.34
&\bf87.18
&77.84\\
\hline
\multicolumn{10}{c}{Korean Kaist (agglutinative)}  \\
\hline
spaCy
&85.21
&72.33
&NA
&NA
&76.15
&68.13
&61.98
&57.52
&NA\\
Stanza
&95.45
&\bf86.31
&NA
&\bf93.02
&88.42
&86.39
&83.97
&80.64
&\bf77.59\\
COMBO
&94.46
&81.66
&NA
&89.16
&87.31
&85.12
&82.70
&78.38
&72.79\\
COMBO$_{\textsc{bert}}$
&\bf95.89
&85.16
&NA
&89.95
&\bf89.77
&\bf87.83
&\bf85.96
&\bf82.66
&75.89\\
\hline
\multicolumn{10}{c}{Turkish IMST (agglutinative)}  \\
\hline
spaCy
&87.66
&86.18
&82.26
&NA
&60.43
&51.32
&47.74
&37.28
&NA\\
Stanza
&\bf95.98
&\bf95.18
&\bf93.77
&96.73
&74.14
&67.52
&64.03
&58.13
&61.91\\
COMBO
&93.60
&92.36
&88.88
&96.47
&72.00
&64.48
&60.48
&49.88
&58.75\\
COMBO$_{\textsc{bert}}$
&95.14
&94.27
&93.56
&\bf97.54
&\bf78.53
&\bf72.03
&\bf68.88
&\bf60.55
&\bf67.13\\
\hline
\multicolumn{10}{c}{Basque BDT (agglutinative with fusional verb morphology)}  \\
\hline
spaCy
&91.96
&NA
&86.67
&NA
&76.11
&70.28
&66.96
&54.46
&NA\\
Stanza
&96.23
&NA
&93.09
&\bf96.52
&86.19
&82.76
&81.30
&73.56
&78.27\\
COMBO
&94.28
&NA
&90.44
&95.47
&84.64
&80.44
&78.82
&67.33
&74.95\\
COMBO$_{\textsc{bert}}$
&\bf96.26
&NA
&\bf93.84
&96.38
&\bf88.73
&\bf85.80
&\bf84.93
&\bf75.96
&\bf81.25\\
\hline
\multicolumn{10}{c}{\bf Average scores}  \\
\hline
spaCy
&91.03
&85.67
&86.97
&NA
&76.73
&70.51
&66.60
&56.57
&NA\\
Stanza
&96.67
&93.96
&94.66
&\bf96.05
&87.01
&83.61
&81.14
&75.60
&77.63\\
COMBO
&95.66
&92.54
&93.05
&94.04
&86.02
&82.21
&79.49
&72.05
&73.98\\
COMBO$_{\textsc{bert}}$
&\bf96.88
&\bf94.37
&\bf95.63
&94.75
&\bf89.67
&\bf86.64
&\bf84.66
&\bf78.92
&\bf79.01\\
\bottomrule
\end{tabular}
\end{center}
}
\caption{\label{tab:2} Processing quality (F$_1$ scores) of spaCy, Stanza and COMBO on the~selected UD treebanks (the language types are given in parentheses). The~highest scores are marked in bold.}
\end{table*}
\section{COMBO Performance}
\label{sec:evaluation}

\paragraph{Data}
COMBO is evaluated on treebanks from the~Universal Dependencies repository \cite{ud25data}, preserving the~original splits into training, validation, and test sets. The~treebanks representing distinctive language types are summarised in Table \ref{tab:statistics} in Appendix~\ref{sec:data_statistics}.

By default, pre-trained 300-dimensional fastText embeddings \cite{grave2018learning} are used. We also test encoding data with pre-trained contextual word embeddings (the~tested BERT models are listed in Table \ref{tab:berts} in Appendix \ref{sec:data_statistics}). 
The~UD datasets provide gold-standard tokenisation. If BERT intra-tokeniser splits a~word into sub-words, the~last layer embeddings are averaged to obtain a~single vector representation of this word.

\paragraph{Qualitative Evaluation}
\label{sec:resultOverview}
Table \ref{tab:2} shows COMBO results of processing the~selected UD treebanks.\footnote{Check the~prediction quality for other languages at: \compacturl{https://gitlab.clarin-pl.eu/syntactic-tools/combo/-/blob/master/docs/performance.md}.} COMBO is compared with Stanza \cite{qi-etal-2020-stanza} and spaCy.\footnote{\compacturl{https://spacy.io}
We use the~project template \compacturl{https://github.com/explosion/projects/tree/v3/pipelines/tagger_parser_ud}. 
The~lemmatiser is implemented as a standalone pipeline component in spaCy v3 and we do not test it.} The~systems are evaluated with the~standard metrics \cite{zeman-EtAl:2018:K18-2}: \textsc{F1}, \textsc{uas} (unlabelled attachment score), \textsc{las} (labelled attachment score), \textsc{mlas} (morphology-aware \textsc{las}) and \textsc{blex} (bi-lexical dependency score).\footnote{\compacturl{http://universaldependencies.org/conll18/conll18_ud_eval.py} (CoNLL 2018 evaluation script).}

COMBO and Stanza undeniably outrun spaCy models.
COMBO using non-contextualised word embeddings is outperformed by Stanza in many language scenarios. However, COMBO supported with BERT-like word embeddings beats all other solutions and is currently the~SOTA system for morphosyntactic analysis.

Regarding lemmatisation, Stanza has an~advantage over COMBO in most tested 
languages. This is probably due to the fact that Stanza lemmatiser is enhanced with a~key-value dictionary, whilst COMBO lemmatiser is fully neural. It is not surprising that a~dictionary helps in lemmatisation of isolating languages (English). However, the~dictionary approach is also helpful for agglutinative languages (Finnish, Korean, Basque) and for Arabic, but not for Polish (fusional languages). 
Comparing COMBO models estimated with and without BERT embeddings, we note that BERT boost only slightly increases the~quality of lemma prediction in the~tested fusional and agglutinative~languages.

\begin{table*}[ht!]
\renewcommand\tabcolsep{9pt}
\setlength\aboverulesep{1pt}
\setlength\belowrulesep{1pt}
\fontsize{10}{12}\selectfont{
\begin{center}
\begin{tabular}{l|c|cccc|cc}
\toprule 
\multirow{2}{*}{Treebank} & spaCy & \multicolumn{4}{c|}{Stanza} & \multicolumn{2}{c}{COMBO}\\
       & & Tagger & Lemmatiser & Parser & Total &  fastText & BERT\\
\midrule
English EWT & 00:22:34 & 02:08:51 & 02:12:17 & 02:29:13 & 06:50:21 & 01:26:55 & 1:54:11 \\
Polish PDB & 01:07:55 & 04:36:51 & 03:19:04 & 05:08:41 & 13:04:36 & 02:39:44 & 3:31:41\\
\bottomrule
\end{tabular}
\end{center}}
\caption{\label{tab:5}Training time of spaCy, Stanza and COMBO.
}
\end{table*}

For a~complete insight into the~prediction quality, we evaluate individual \textsc{upos} and \textsc{udeprel} predictions in English (the~isolating language), Korean (agglutinative) and Polish (fusional). Result visualisations are in Appendix \ref{sec:visualisations}.

COMBO took part in IWPT 2021 Shared Task on Parsing into Enhanced Universal Dependencies \cite{bouma-etal-2021-raw}, 
where it ranked  4th.\footnote{\compacturll{https://universaldependencies.org/iwpt21/results.html}} In addition to \textsc{elas} and \textsc{eulas} metrics, the~third evaluation metric was \textsc{las}. COMBO ranked 2nd, achieving the~average \textsc{las} of 87.84\%. The~score is even higher than the~average \textsc{las} of 86.64\% in Table \ref{tab:2}, which is a~kind of confirmation that our evaluation is representative, reliable, and fair.

\paragraph{Downstream Evaluation}
According to the~results in Table \ref{tab:2}, COMBO predicts high-quality dependency trees and parts of speech. We therefore conduct a~preliminary evaluation of morphosyntactically informed word embeddings in the~textual entailment task (aka natural language inference, NLI) in English \cite{Bentivogli:2016} and Polish \cite{wroblewska-krasnowska-kieras-2017-polish}.
We compare the~quality of entailment classifiers with two FC layers trained on max/mean-pooled BERT embeddings and sentence representations estimated by a~network with two transformer layers which is given morphosyntactically informed word embeddings (i.e. BERT-based word embeddings concatenated with \textsc{upos} embeddings, \textsc{deprel} embeddings, and BERT-based embeddings of the~head word). The~morphosyntactically informed English NLI classifier achieves an~accuracy of 78.84\% and outperforms the~max/mean-pooled classifiers by 20.77~pp and 5.44~pp, respectively. The~Polish syntax-aware NLI classifier achieves an~accuracy of 91.60\% and outperforms the~max/mean-pooled classifiers by 17.2~pp and 7.7~pp, respectively.

\paragraph{Efficiency Evaluation}
We also compare spaCy, Stanza and COMBO in terms of their efficiency, i.e. training and prediction speed.\footnote{\label{footnote:gpu}A~single NVIDIA V100 card is used in all tests.} 
According to the~results (see Tables \ref{tab:5} and \ref{tab:6}), spaCy is the~SOTA system, and the~other two are not even close to its processing time. Considering COMBO and Stanza, whose prediction quality is significantly better than spaCy, COMBO is 1.5 times slower (2 times slower with BERT) than Stanza in predicting, but it is definitely faster in training. The~reason for large discrepancies in training times is the~different architecture of these two systems. Stanza is a~pipeline-based system, i.e. its modules are trained one after the~other. COMBO is an~end-to-end system, i.e. its modules are jointly trained and the~training process is therefore faster.

\begin{table}[ht!]
\setlength\aboverulesep{1pt}
\setlength\belowrulesep{1pt}
\fontsize{10}{12}\selectfont{
\begin{center}
\begin{tabular*}{\linewidth}{l|ccc}
\toprule 
Treebank & Stanza & COMBO & COMBO$_{\textsc{bert}}$\\
\midrule
English EWT & $4.7\times$ & $6.8\times$ & $10.8\times$\\
Polish PDB & $4.1\times$ & $5.8\times$ & $10.6\times$\\
\bottomrule
\end{tabular*}
\end{center}}
\caption{\label{tab:6}Prediction time of Stanza and COMBO relative to spaCy ($1\times$) on 
English and Polish test data.
}
\end{table}

\section{Getting Started with COMBO}
\label{sec:getting_started}
\paragraph{Prediction} COMBO provides two main prediction modes: a~Python library and a~command-line interface (CLI). The~Python package mode supports automated model download. The~code snippet demonstrates downloading a~pre-trained Polish model and processing a~sentence:
\vspace{-0.5cm}
\begin{minted}
[
frame=lines,
framesep=2mm,
baselinestretch=1.2,
fontsize=\footnotesize,
]
{python}
from combo.predict import COMBO

nlp = COMBO.from_pretrained("polish")
sentence = nlp("Ala ma kota.")
print(sentence.tokens)
\end{minted}
\vspace{-0.4cm}
To download a~model for another language, select its name from the~list of pre-trained models.\footnote{The~list of the~pretrained COMBO models:  \compacturll{https://gitlab.clarin-pl.eu/syntactic-tools/combo/-/blob/master/docs/models.md\#pre-trained-models}} The~Python mode also supports acquisition of \textsc{deprel} or \textsc{upos} embeddings, for example:
\vspace{-0.5cm}
\begin{minted}
[
frame=lines,
framesep=2mm,
baselinestretch=1.2,
fontsize=\footnotesize,
]
{python}
sentence = nlp("Ala ma kota.")
chosen_token = sentence.tokens[1]
print(chosen_token.embeddings["upostag"])
\end{minted}
\vspace{-0.4cm}
In CLI mode, COMBO processes sentences using either a~downloaded model or a~model trained by yourself. 
CLI works on raw texts and on the~CoNLL-U files (i.e. with tokenised sentences and even morphologically annotated tokens):
\vspace{-0.5cm}
\begin{minted}
[
frame=lines,
framesep=2mm,
baselinestretch=1.2,
fontsize=\footnotesize,
]
{bash}
combo --mode predict \
      --model_path model.tar.gz \
      --input_file input.conllu \
      --output_file output.conllu
\end{minted}
\vspace{-0.4cm}

\paragraph{Model Training}
COMBO CLI allows to train new models for any language. The~only requirement is a~training dataset in the~CoNLL-U/CoNLL-X 
format. In the~default setup, tokenised sentences are input and all possible predictors are trained:
\vspace{-0.5cm}
\begin{minted}
[
frame=lines,
framesep=2mm,
baselinestretch=1.2,
fontsize=\footnotesize,
]
{bash}
combo --mode train \
      --training_data training.conllu \
      --validation_data valid.conllu
\end{minted}
\vspace{-0.4cm}

\noindent
If we only train a~dependency parser, the~default setup should be changed with configuration flags: \mintinline{bash}{--features} with a~list of input features and \mintinline{bash}{--targets} with a~list of prediction targets.

\section{Conclusion}
\label{sec:conclusions}

We have presented COMBO, the~SOTA system for morphosyntacic analysis, i.e. part-of-speech tagging, morphological analysis, lemmatisation, and (enhanced) dependency parsing. COMBO is a~language-agnostic and format-independent system (i.e. it supports the~CoNLL-U and CoNLL-X formats). Its implementation as a Python package allows effortless installation, and incorporation into any Python code or usage in the~CLI mode. In the~Python mode, COMBO supports automated download of pre-trained models for multiple languages and outputs not only categorical morphosyntactic features, but also their embeddings. In the~CLI mode, pre-trained models can be manually downloaded or trained from scratch. The~system training is fully configurable in respect of the~range of input features and output predictions, and the~method of encoding input data.

\noindent
Last but not least, COMBO maintains a balance between efficiency and quality. Admittedly, it is not as fast as spaCy, but it is much more efficient than Stanza considering the~training time. Tested on the~selected UD treebanks, COMBO morphosyntactic models enhanced with BERT embeddings outperform spaCy and Stanza models. 

\section*{Acknowledgments}
The authors would like to thank Piotr Rybak for his design and explanations of the~architecture of COMBO's forerunner. The research presented in this paper was founded by SONATA 8 grant no 2014/15/D/HS2/03486 from the National Science Centre Poland and the European Regional Development Fund as a part of the 2014-2020 Smart Growth Operational Programme, CLARIN -- Common Language Resources and Technology Infrastructure, project no. POIR.04.02.00-00C002/19. The~computing was performed at Pozna\'{n} Supercomputing and Networking Center.

\bibliography{custom,anthology}
\bibliographystyle{acl_natbib}

\clearpage
\appendix

\section{COMBO Implementation}
\label{sec:implementationDetails}
COMBO is a~Python package that uses the~PyTorch \cite{NEURIPS2019_9015} and AllenNLP \cite{Gardner2017AllenNLP} libraries. 
The~COMBO models used in the~evaluation presented in Section~\ref{sec:evaluation} are trained with the~empirically set default parameters specified below.
The~training parameters can be easily configured and adjusted to the~specifics of an~individual model.

\subsection{Network Hyperparameters}
\label{sec:hyperparameters}

\paragraph{Embeddings}
An~internal character-based word embedding is calculated with three convolutional layers with 512, 256 and 64 filters with dilation rates equal to 1, 2 and 4. All filters have the~kernel size of 3. The~internal word embedding has a~size of 64 dimensions.
All external word embeddings are reduced to 100-dimensional vectors by a~single FC layer.
As only words are used as input features in the~system evaluation, the~local feature embedding is a~concatenation of the~64-dimensional internal and 100-dimensional external word embedding.
The~global feature vectors are computed by two bi\-LSTM layers with 512 hidden units.

\paragraph{Prediction modules}
The tagger uses a~FC network with a~hidden layer of the~size 64 to predict \textsc{upos} and FC networks with 128-dimensional hidden layers to predict \textsc{xpos} and \textsc{ufeats}.

\noindent
The lemmatiser uses three convolutional layers with 256 filters and dilation rates equal to 1, 2 and 4. All filters have the kernel size of 3. The~fourth convolutional layer with the~number of filters equal to the~number of character instances in training data is used to predict the~probability of each character. The~final layer filters have the~kernel size of 1. The 256-dimensional embeddings of input characters are concatenated with
the~global feature vectors reduced to 32 dimensions with a~single FC layer.

\noindent
The~arc prediction module uses 512-dimensional head, and dependent embeddings and the~labelling module uses 128-dimensional vectors. 

COMBO-vectoriser currently outputs 64-dimensional \textsc{upos} and 128-dimensional \textsc{deprel} embeddings.

\paragraph*{Activation function} 
FC and CNN layers use hyperbolic tangent and rectified linear unit \cite{relu:2010} activation functions, respectively.

\subsection{Regularisation}
Dropout technique for Variational RNNs \cite{variational_rnn} with 0.33 rate is applied to the~local feature embeddings and on top of the~stacked biLSTM estimating global feature embeddings. The~same dropout, for output and recurrent values, is used in the context of each biLSTM layer.
The~FC layers use the~standard dropout \cite{dropout} with 0.25 rate.
Moreover, the~biLSTM and convolutional layers use L2 regularisation with the~rate of $1 \times 10^{-6}$, and the~trainable embeddings use L2 with the~rate of $1 \times 10^{-5}$.

\subsection{Training}
The~cross-entropy loss is used for all parts of the~system. The~final loss is the~weighted sum of losses with the~following weights for each task:
\begin{itemize}
    \itemsep-.4em 
    \item 0.05 for predicting \textsc{upos} and \textsc{lemma},
    \item 0.2 for predicting \textsc{ufeats} and (enh)\textsc{head},
    \item 0.8 for predicting (enh)\textsc{deprel}.
\end{itemize}

\noindent
The~whole system is optimised with ADAM \cite{Kingma:2014} with the~learning rate of $0.002$ and $\beta_1 = \beta_2 = 0.9$. 
The~model is trained for a~maximum of 400 epochs, and the~learning rate is reduced twice by the~factor of two when the~validation score reaches a~plateau.

\section{External Data Summary}
\label{sec:data_statistics}
Tables \ref{tab:statistics} and \ref{tab:berts} list the~UD dependency treebanks and BERT models used in the~evaluation experiments presented in Section \ref{sec:evaluation}.

\begin{table*}[ht!]
\centering
\renewcommand\tabcolsep{6pt}
\renewcommand{\arraystretch}{0.9}
\setlength\aboverulesep{1pt}
\setlength\belowrulesep{1pt}
\fontsize{10}{12}\selectfont{
\begin{tabular}{lllrrl}
\toprule
Language & Language Type & UD Treebank  & \#Words & \#Trees & Reference \\
\midrule
English & isolating & English-EWT  & 254,856 & 16,622 & \citet{silveira-etal-2014-gold}\\
Arabic & fusional & Arabic-PADT  & 282,384 & 7,664 & \citet{arabic-UD}\\
Polish & fusional & Polish-PDB & 350,036 & 22,152 & \citet{wrob:18} \\
Finnish & agglutinative & Finnish-TDT & 202,453 & 15,136 & \citet{finnish-UD}\\
Korean & agglutinative & Korean-Kaist  & 350,090 & 27,363 & \citet{korean-UD}\\
Turkish & agglutinative & Turkish-IMST & 57,859& 5,635 & \citet{turkish-UD}\\
Basque & agglutinative (fusional & Basque-BDT & 121,443 & 8,993 & \citet{basque-UD} \\
& verb morphology) & & & \\
\bottomrule
\end{tabular}}
\caption{\label{tab:statistics} The UD treebanks used in the evaluation experiments.}
\end{table*}

\begin{table*}[ht!]
\centering
\renewcommand\tabcolsep{6pt}
\renewcommand{\arraystretch}{0.9}
\setlength\aboverulesep{1pt}
\setlength\belowrulesep{1pt}
\fontsize{10}{12}\selectfont{
\begin{tabular}{lll}
\toprule
Language  & BERT model & Reference\\
\midrule
Arabic & bert-base-arabertv2 & \citet{antoun-etal-2020-arabert} \\
Basque  & berteus-base-cased & \citet{agerri-etal-2020-give} \\
English  & bert-base-cased & \citet{bert:2018}\\
Finnish  & bert-base-finnish-cased-v1 & \citet{virtanen2019multilingual}\\
Korean & bert-kor-base &  \citet{kim2020lmkor}\\
Polish  & herbert-base-cased & \citet{mroczkowski-etal-2021-herbert}\\
Turkish & bert-base-turkish-cased & \citet{stefan_schweter_2020_3770924}\\
\bottomrule
\end{tabular}}
\caption{\label{tab:berts} The BERT models used in the~evaluation experiments.}
\end{table*}

\section{Evaluation of \textsc{upos} and \textsc{udeprel}}
\label{sec:visualisations}

\begin{figure*}[!h]
\centering
\begin{tikzpicture}
\begin{axis}[
width=\textwidth, height=4.5cm, enlargelimits=0.02, legend style={at={(0.33,1.1)},anchor=west,legend columns=3}, ybar interval=0.7,
ymin=0, ymax=100,
y tick label style={font=\small}, ytick distance=20,
xtick=data, xticklabels={\rotatebox{90}{\small{ADJ}},\rotatebox{90}{\small{ADP}},\rotatebox{90}{\small{ADV}},\rotatebox{90}{\small{AUX}},\rotatebox{90}{\small{CCONJ}},\rotatebox{90}{\small{DET}},\rotatebox{90}{\small{INTJ}},\rotatebox{90}{\small{NOUN}},\rotatebox{90}{\small{NUM}},\rotatebox{90}{\small{PART}},\rotatebox{90}{\small{PRON}},\rotatebox{90}{\small{PROPN}},\rotatebox{90}{\small{PUNCT}},\rotatebox{90}{\small{SCONJ}},\rotatebox{90}{\small{SYM}},\rotatebox{90}{\small{VERB}},\rotatebox{90}{\small{X}},\rotatebox{90}{\small{}}},
x tick label style={rotate=-45, anchor=north, align=right},]
\addplot coordinates {(0,86) (1,96) (2,79) (3,70) (4,88) (5,88) (6,24) (7,81) (8,86) (9,80) (10,90) (11,71) (12,84) (13,87) (14,25) (15,85) (16,57) (17,0) };
\addplot coordinates {(0,94) (1,97) (2,87) (3,95) (4,93) (5,95) (6,47) (7,90) (8,91) (9,84) (10,95) (11,89) (12,93) (13,92) (14,22) (15,93) (16,78) (17,0) };
\addplot coordinates {(0,97) (1,98) (2,91) (3,98) (4,96) (5,97) (6,80) (7,94) (8,94) (9,85) (10,97) (11,88) (12,95) (13,96) (14,22) (15,95) (16,80) (17,0) };
\legend{\small{spaCy},\small{stanza},
\small{COMBO$_{\small\textsc{bert}}$}}
\end{axis}
\end{tikzpicture}
\vspace*{-.3cm}
\caption{Evaluation of predicted universal parts of speech (\textsc{upos}) in the~Polish test set (F-$_{1}$-scores).}
\label{fig:pos_pl}
\end{figure*}
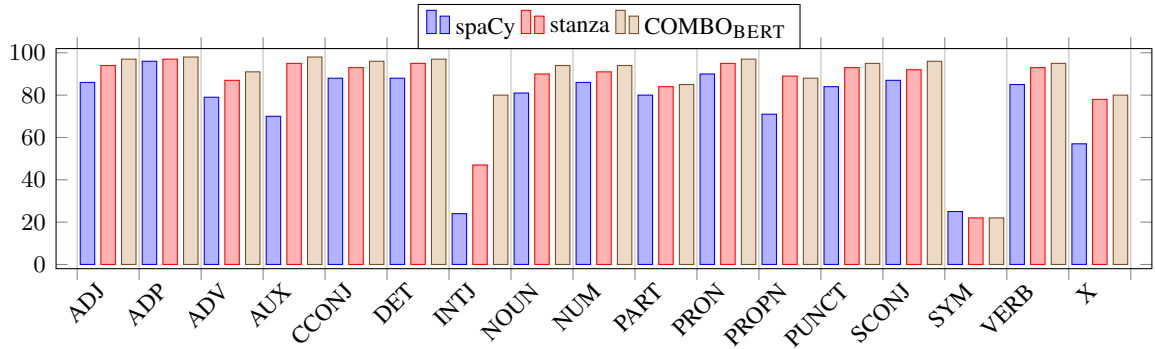

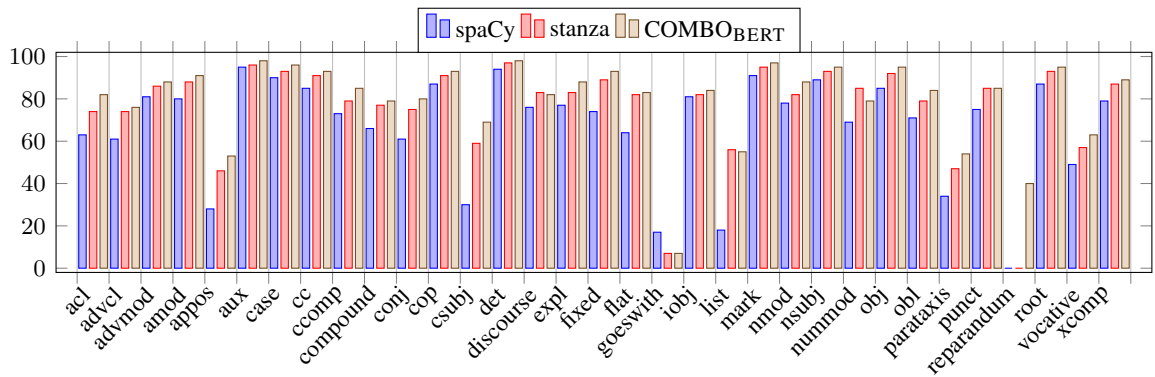
\begin{figure*}[!h]
\centering
\begin{tikzpicture}
\begin{axis}[
width=\textwidth, height=4.5cm, enlargelimits=0.02, legend style={at={(0.33,1.1)},anchor=west,legend columns=3}, ybar interval=0.7,
ymin=0, ymax=100,
y tick label style={font=\small}, ytick distance=20,
xtick=data, xticklabels={\rotatebox{90}{\small{acl}},\rotatebox{90}{\small{advcl}},\rotatebox{90}{\small{advmod}},\rotatebox{90}{\small{amod}},\rotatebox{90}{\small{appos}},\rotatebox{90}{\small{aux}},\rotatebox{90}{\small{case}},\rotatebox{90}{\small{cc}},\rotatebox{90}{\small{ccomp}},\rotatebox{90}{\small{compound}},\rotatebox{90}{\small{conj}},\rotatebox{90}{\small{cop}},\rotatebox{90}{\small{csubj}},\rotatebox{90}{\small{det}},\rotatebox{90}{\small{discourse}},\rotatebox{90}{\small{expl}},\rotatebox{90}{\small{fixed}},\rotatebox{90}{\small{flat}},\rotatebox{90}{\small{goeswith}},\rotatebox{90}{\small{iobj}},\rotatebox{90}{\small{list}},\rotatebox{90}{\small{mark}},\rotatebox{90}{\small{nmod}},\rotatebox{90}{\small{nsubj}},\rotatebox{90}{\small{nummod}},\rotatebox{90}{\small{obj}},\rotatebox{90}{\small{obl}},\rotatebox{90}{\small{parataxis}},\rotatebox{90}{\small{punct}},\rotatebox{90}{\small{reparandum}},\rotatebox{90}{\small{root}},\rotatebox{90}{\small{vocative}},\rotatebox{90}{\small{xcomp}},\rotatebox{90}{\small{}}},
x tick label style={rotate=-45, anchor=north, align=right},]
\addplot coordinates {(0,63) (1,61) (2,81) (3,80) (4,28) (5,95) (6,90) (7,85) (8,73) (9,66) (10,61) (11,87) (12,30) (13,94) (14,76) (15,77) (16,74) (17,64) (18,17) (19,81) (20,18) (21,91) (22,78) (23,89) (24,69) (25,85) (26,71) (27,34) (28,75) (29,0) (30,87) (31,49) (32,79) (33,0)};
\addplot coordinates {(0,74) (1,74) (2,86) (3,88) (4,46) (5,96) (6,93) (7,91) (8,79) (9,77) (10,75) (11,91) (12,59) (13,97) (14,83) (15,83) (16,89) (17,82) (18,7) (19,82) (20,56) (21,95) (22,82) (23,93) (24,85) (25,92) (26,79) (27,47) (28,85) (29,0) (30,93) (31,57) (32,87) (33,0)};
\addplot coordinates {(0,82) (1,76) (2,88) (3,91) (4,53) (5,98) (6,96) (7,93) (8,85) (9,79) (10,80) (11,93) (12,69) (13,98) (14,82) (15,88) (16,93) (17,83) (18,7) (19,84) (20,55) (21,97) (22,88) (23,95) (24,79) (25,95) (26,84) (27,54) (28,85) (29,40) (30,95) (31,63) (32,89) (33,0)};
\legend{\small{spaCy},\small{stanza},
\small{COMBO$_{\small\textsc{bert}}$}}
\end{axis}
\end{tikzpicture}
\vspace*{-.3cm}
\caption{\label{fig:gf}Evaluation of predicted grammatical functions (\textsc{udeprel}) in the~English test set (F-$_{1}$-scores).
}
\end{figure*}

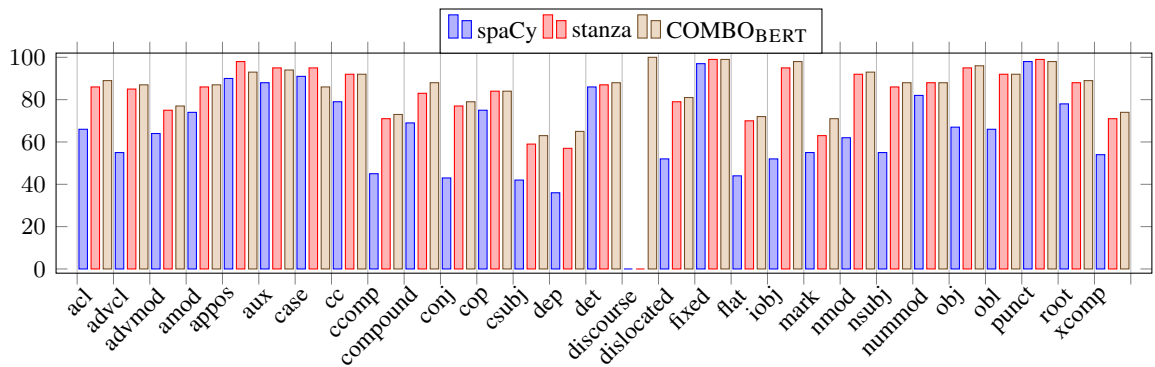
\begin{figure*}[!h]
\centering
\begin{tikzpicture}
\begin{axis}[
width=\textwidth, height=4.5cm, enlargelimits=0.02, legend style={at={(0.35,1.1)},anchor=west,legend columns=3}, ybar interval=0.7,
ymin=0, ymax=100,
y tick label style={font=\small}, ytick distance=20,
xtick=data, xticklabels={\rotatebox{90}{\small{acl}},
\rotatebox{90}{\small{advcl}},
\rotatebox{90}{\small{advmod}},
\rotatebox{90}{\small{amod}},
\rotatebox{90}{\small{appos}},
\rotatebox{90}{\small{aux}},
\rotatebox{90}{\small{case}},
\rotatebox{90}{\small{cc}},
\rotatebox{90}{\small{ccomp}},
\rotatebox{90}{\small{compound}},
\rotatebox{90}{\small{conj}},
\rotatebox{90}{\small{cop}},
\rotatebox{90}{\small{csubj}},
\rotatebox{90}{\small{dep}},
\rotatebox{90}{\small{det}},
\rotatebox{90}{\small{discourse}},
\rotatebox{90}{\small{dislocated}},
\rotatebox{90}{\small{fixed}},
\rotatebox{90}{\small{flat}},
\rotatebox{90}{\small{iobj}},
\rotatebox{90}{\small{mark}},
\rotatebox{90}{\small{nmod}},
\rotatebox{90}{\small{nsubj}},
\rotatebox{90}{\small{nummod}},
\rotatebox{90}{\small{obj}},
\rotatebox{90}{\small{obl}},
\rotatebox{90}{\small{punct}},
\rotatebox{90}{\small{root}},
\rotatebox{90}{\small{xcomp}},
\rotatebox{90}{\small{}}},
x tick label style={rotate=-45, anchor=north, align=right},]
\addplot coordinates {(0,66) (1,55) (2,64) (3,74) (4,90) (5,88) (6,91) (7,79) (8,45) (9,69) (10,43) (11,75) (12,42) (13,36) (14,86) (15,00) (16,52) (17,97) (18,44) (19,52) (20,55) (21,62) (22,55) (23,82) (24,67) (25,66) (26,98) (27,78) (28,54) (29,0)};
\addplot coordinates {(0,86) (1,85) (2,75) (3,86) (4,98) (5,95) (6,95) (7,92) (8,71) (9,83) (10,77) (11,84) (12,59) (13,57) (14,87) (15,00) (16,79) (17,99) (18,70) (19,95) (20,63) (21,92) (22,86) (23,88) (24,95) (25,92) (26,99) (27,88) (28,71) (29,0)};
\addplot coordinates {(0,89) (1,87) (2,77) (3,87) (4,93) (5,94) (6,86) (7,92) (8,73) (9,88) (10,79) (11,84) (12,63) (13,65) (14,88) (15,100) (16,81) (17,99) (18,72) (19,98) (20,71) (21,93) (22,88) (23,88) (24,96) (25,92) (26,98) (27,89) (28,74) (29,0)};
\legend{\small{spaCy},\small{stanza},
\small{COMBO$_{\small\textsc{bert}}$}}
\end{axis}
\end{tikzpicture}
\vspace*{-.3cm}
\caption{Evaluation of predicted grammatical functions (\textsc{udeprel}) in the~Korean test set (F-$_{1}$-scores).}
\label{fig:gf_ko}
\end{figure*}

\begin{figure*}[!h]
\centering
\begin{tikzpicture}
\begin{axis}[
width=\textwidth, height=4.5cm, enlargelimits=0.02, legend style={at={(0.33,1.1)},anchor=west,legend columns=3}, ybar interval=0.7,
ymin=0, ymax=100,
y tick label style={font=\small}, ytick distance=20,
xtick=data, xticklabels={\rotatebox{90}{\small{acl}},\rotatebox{90}{\small{advcl}},\rotatebox{90}{\small{advmod}},\rotatebox{90}{\small{amod}},\rotatebox{90}{\small{appos}},\rotatebox{90}{\small{aux}},\rotatebox{90}{\small{case}},\rotatebox{90}{\small{cc}},\rotatebox{90}{\small{ccomp}},\rotatebox{90}{\small{conj}},\rotatebox{90}{\small{cop}},\rotatebox{90}{\small{csubj}},\rotatebox{90}{\small{det}},\rotatebox{90}{\small{discourse}},\rotatebox{90}{\small{expl}},\rotatebox{90}{\small{fixed}},\rotatebox{90}{\small{flat}},\rotatebox{90}{\small{iobj}},\rotatebox{90}{\small{list}},\rotatebox{90}{\small{mark}},\rotatebox{90}{\small{nmod}},\rotatebox{90}{\small{nsubj}},\rotatebox{90}{\small{nummod}},\rotatebox{90}{\small{obj}},\rotatebox{90}{\small{obl}},\rotatebox{90}{\small{orphan}},\rotatebox{90}{\small{parataxis}},\rotatebox{90}{\small{punct}},\rotatebox{90}{\small{root}},\rotatebox{90}{\small{vocative}},\rotatebox{90}{\small{xcomp}},\rotatebox{90}{\small{}}},
x tick label style={rotate=-45, anchor=north, align=right},]
\addplot coordinates {(0,67) (1,68) (2,80) (3,90) (4,54) (5,57) (6,96) (7,89) (8,71) (9,67) (10,79) (11,44) (12,90) (13,43) (14,96) (15,81) (16,76) (17,68) (18,48) (19,87) (20,74) (21,81) (22,90) (23,77) (24,79) (25,25) (26,56) (27,84) (28,92) (29,55) (30,87) (31,0)};
\addplot coordinates {(0,83) (1,79) (2,86) (3,96) (4,75) (5,95) (6,98) (7,94) (8,86) (9,82) (10,89) (11,59) (12,96) (13,53) (14,99) (15,85) (16,93) (17,83) (18,87) (19,92) (20,84) (21,92) (22,95) (23,91) (24,88) (25,0) (26,71) (27,93) (28,97) (29,85) (30,92) (31,0)};
\addplot coordinates {(0,90) (1,85) (2,88) (3,97) (4,81) (5,98) (6,99) (7,96) (8,90) (9,89) (10,97) (11,81) (12,98) (13,74) (14,98) (15,87) (16,92) (17,89) (18,96) (19,96) (20,89) (21,97) (22,96) (23,95) (24,91) (25,44) (26,74) (27,95) (28,99) (29,87) (30,95) (31,0)};
\legend{\small{spaCy},\small{stanza},
\small{COMBO$_{\small\textsc{bert}}$}}
\end{axis}
\end{tikzpicture}
\vspace*{-.3cm}
\caption{\label{fig:gf_pl}Evaluation of predicted grammatical functions (\textsc{udeprel}) in the~Polish test set (F-$_{1}$-scores).}
\end{figure*}
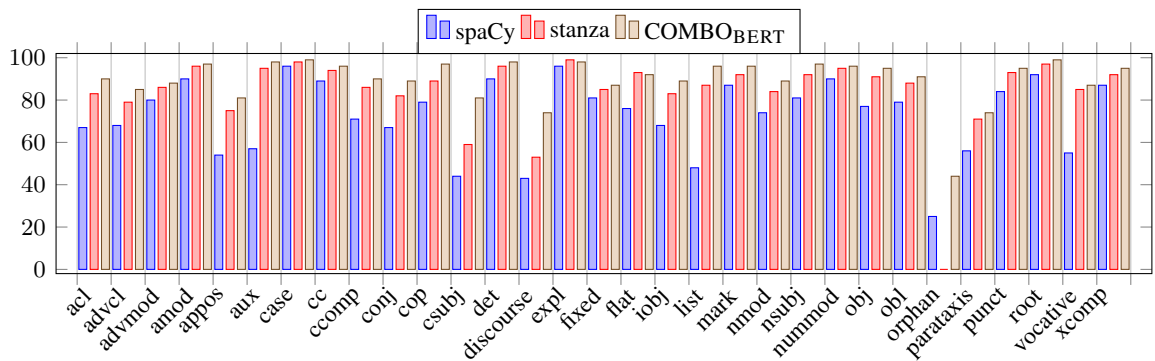

The~comparison of the~universal parts of speech predicted by the tested systems in English, Korean and Polish data is shown in the~charts in Figures \ref{fig:pos}, \ref{fig:pos_ko} and \ref{fig:pos_pl}, respectively. 
The~comparison of the~quality of the~predicted universal dependency types in English, Korean and Polish data is presented in Figures \ref{fig:gf}, \ref{fig:gf_ko} and \ref{fig:gf_pl}, respectively. 

\begin{figure*}[!h]
\centering
\begin{tikzpicture}
\begin{axis}[
width=\textwidth, height=4.5cm, enlargelimits=0.02, legend style={at={(0.33,1.1)},anchor=west,legend columns=3}, ybar interval=0.7,
ymin=0, ymax=100,
y tick label style={font=\small}, ytick distance=20,
xtick=data, xticklabels={\rotatebox{90}{\small{ADJ}},\rotatebox{90}{\small{ADP}},\rotatebox{90}{\small{ADV}},\rotatebox{90}{\small{AUX}},\rotatebox{90}{\small{CCONJ}},\rotatebox{90}{\small{DET}},\rotatebox{90}{\small{INTJ}},\rotatebox{90}{\small{NOUN}},\rotatebox{90}{\small{NUM}},\rotatebox{90}{\small{PART}},\rotatebox{90}{\small{PRON}},\rotatebox{90}{\small{PROPN}},\rotatebox{90}{\small{PUNCT}},\rotatebox{90}{\small{SCONJ}},\rotatebox{90}{\small{SYM}},\rotatebox{90}{\small{VERB}},\rotatebox{90}{\small{X}},\rotatebox{90}{\small{}}},
x tick label style={rotate=-45, anchor=north, align=right},]
\addplot coordinates {(0,79) (1,88) (2,79) (3,93) (4,87) (5,93) (6,76) (7,73) (8,65) (9,94) (10,93) (11,61) (12,75) (13,80) (14,46) (15,78) (16,44) (17,0)};
\addplot coordinates {(0,87) (1,92) (2,85) (3,95) (4,93) (5,97) (6,85) (7,83) (8,81) (9,96) (10,95) (11,75) (12,85) (13,90) (14,60) (15,86) (16,61) (17,0)};
\addplot coordinates {(0,88) (1,95) (2,86) (3,96) (4,95) (5,98) (6,81) (7,86) (8,80) (9,97) (10,96) (11,78) (12,85) (13,94) (14,61) (15,90) (16,55) (17,0)};
\legend{\small{spaCy},\small{stanza},
\small{COMBO$_{\small\textsc{bert}}$}}
\end{axis}
\end{tikzpicture}
\vspace*{-.3cm}
\caption{Evaluation of predicted universal parts of speech (\textsc{upos}) in the~English test set (F-$_{1}$-scores).}
\label{fig:pos}
\end{figure*}
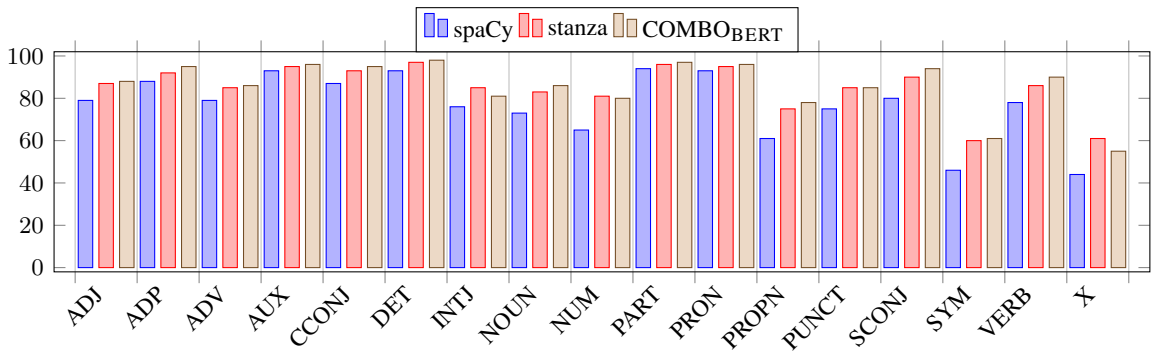

\begin{figure*}[!h]
\centering
\begin{tikzpicture}
\begin{axis}[
width=\textwidth, height=4.5cm, enlargelimits=0.02, legend style={at={(0.33,1.1)},anchor=west,legend columns=3}, ybar interval=0.7,
ymin=0, ymax=100,
y tick label style={font=\small}, ytick distance=20,
xtick=data, xticklabels={\rotatebox{90}{\small{ADJ}},
\rotatebox{90}{\small{ADP}},
\rotatebox{90}{\small{ADV}},
\rotatebox{90}{\small{AUX}},
\rotatebox{90}{\small{CCONJ}},
\rotatebox{90}{\small{DET}},
\rotatebox{90}{\small{INTJ}},
\rotatebox{90}{\small{NOUN}},
\rotatebox{90}{\small{NUM}},
\rotatebox{90}{\small{PRON}},
\rotatebox{90}{\small{PROPN}},
\rotatebox{90}{\small{PUNCT}},
\rotatebox{90}{\small{SCONJ}},
\rotatebox{90}{\small{SYM}},
\rotatebox{90}{\small{VERB}},
\rotatebox{90}{\small{X}},
\rotatebox{90}{\small{}}},
x tick label style={rotate=-45, anchor=north, align=right},]
\addplot coordinates {(0,73) (1,88) (2,65) (3,78) (4,49) (5,87) (6,00) (7,65) (8,73) (9,75) (10,52) (11,98) (12,50) (13,96) (14,69) (15,78) (16,00)};
\addplot coordinates {(0,83) (1,97) (2,85) (3,91) (4,79) (5,87) (6,00) (7,85) (8,87) (9,86) (10,73) (11,99) (12,74) (13,96) (14,84) (15,94) (16,00)};
\addplot coordinates {(0,84) (1,88) (2,87) (3,89) (4,82) (5,88) (6,50) (7,88) (8,88) (9,84) (10,78) (11,98) (12,76) (13,87) (14,86) (15,85) (16,00)};
\legend{\small{spaCy},\small{stanza},
\small{COMBO$_{\small\textsc{bert}}$}}
\end{axis}
\end{tikzpicture}
\vspace*{-.3cm}
\caption{Evaluation of predicted universal parts of speech (\textsc{upos}) in the~Korean test set (F-$_{1}$-scores).}
\label{fig:pos_ko}
\end{figure*}
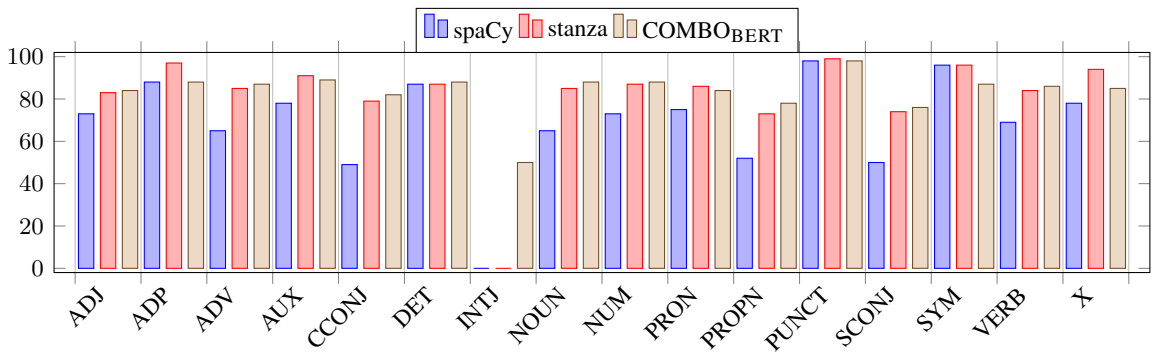

\end{document}